\documentclass[conference]{IEEEtran}
\IEEEoverridecommandlockouts
\usepackage{cite}
\usepackage{amsmath,amssymb,amsfonts}
\usepackage{algorithmic}
\usepackage{graphicx}
\usepackage{textcomp}
\usepackage{xcolor}
\usepackage{url}

\usepackage{enumitem}
\usepackage{tcolorbox}
\tcbuselibrary{listingsutf8}
\usepackage{listings}

\usepackage{multirow}
\usepackage{booktabs}

\usepackage{tabularx}
\usepackage{array}
\usepackage{booktabs}
\usepackage{makecell}
\usepackage{hyperref}

\definecolor{lightblueframe}{HTML}{99CCFF}

\usepackage{soul}
\usepackage{wrapfig}
\usepackage{subcaption}

\definecolor{mid-blue}{HTML}{3399CC}

\usepackage{amsthm}

\theoremstyle{problemstyle}
\newtheorem{problem}{Problem}

\def\BibTeX{{\rm B\kern-.05em{\sc i\kern-.025em b}\kern-.08em
    T\kern-.1667em\lower.7ex\hbox{E}\kern-.125emX}}
\begin{document}

\title{Optimizing EEG Graph Structure for Seizure Detection: An Information Bottleneck and Self-Supervised Learning Approach}

% \author{\IEEEauthorblockN{Lincan Li}
% \IEEEauthorblockA{\textit{Department of Computer Science} \\
% \textit{Florida State University}\\
% Tallahassee, USA \\
% %ll24bb@fsu.edu
% }
% \and
% \IEEEauthorblockN{Rikuto Kotoge}
% \IEEEauthorblockA{\textit{SANKEN} \\
% \textit{The University of Osaka}\\
% Ibaraki, Japan \\
% %rikuto88@sanken.osaka-u.ac.jp
% }
% \and
% \IEEEauthorblockN{Xihao Piao}
% \IEEEauthorblockA{\textit{SANKEN} \\
% \textit{The University of Osaka}\\
% Ibaraki, Japan \\
% %park88@sanken.osaka-u.ac.jp
% }
% \and
% \IEEEauthorblockN{Zheng Chen}
% \IEEEauthorblockA{\textit{SANKEN} \\
% \textit{The University of Osaka}\\
% Ibaraki, Japan \\
% %chenz@sanken.osaka-u.ac.jp
% }
% \and
% \IEEEauthorblockN{Yushun Dong}
% \IEEEauthorblockA{\textit{Department of Computer Science} \\
% \textit{Florida State University}\\
% Tallahassee, USA \\
% %yushun.dong@fsu.edu
% }
% }

\author{
\IEEEauthorblockN{Lincan Li\IEEEauthorrefmark{1},
Rikuto Kotoge\IEEEauthorrefmark{2}, Xihao Piao\IEEEauthorrefmark{2}, Zheng Chen\IEEEauthorrefmark{2}, Yushun Dong\IEEEauthorrefmark{1}\thanks{Corresponding author: Yushun Dong (Email: yushun.dong@fsu.edu).}}
\IEEEauthorblockA{\IEEEauthorrefmark{1}Department of Computer Science, Florida State University, Tallahassee, USA\\
\IEEEauthorrefmark{2}SANKEN, The University of Osaka, Ibaraki, Japan}
}

\maketitle

\begin{abstract}
Seizure detection based on EEG signals is highly challenging due to complex spatiotemporal dynamics and extreme inter-patient variability. To model such complex patterns, recent methods construct dynamic graphs via statistical correlations, predefined similarity measures, or implicit learning, yet rarely account for EEG's highly noisy nature. Consequently, these graphs usually contain redundant or task-irrelevant connections, undermining model performance even when using state-of-the-art architectures. In this paper, we present a new perspective for EEG seizure detection: jointly learning denoised dynamic graph structures and informative spatial-temporal representations guided by the Information Bottleneck (IB). Unlike prior approaches, our graph constructor explicitly accounts for the noisy characteristics of EEG data, producing compact and reliable connectivity patterns that better support downstream seizure detection. To further enhance representation learning, we employ a self-supervised Graph Masked AutoEncoder that reconstructs masked EEG signals based on dynamic graph context, promoting structure-aware and compact representations that align with the IB principle. Bringing things together, we introduce \underline{\textbf{I}}nformation Bottleneck-guided EEG Seizu\underline{\textbf{RE}} Detectio\underline{\textbf{N}} via S\underline{\textbf{E}}lf-Supervised Learning (\textbf{IRENE}), which explicitly learns dynamic graph structures and interpretable spatial-temporal EEG representations. IRENE addresses three core challenges: \textbf{(i)} Identifying the most informative nodes and edges; \textbf{(ii)} Explaining seizure propagation in the brain network; and \textbf{(iii)} Enhancing robustness against label scarcity and inter-patient variability. Extensive experiments on benchmark EEG datasets demonstrate that our method outperforms state-of-the-art baselines in seizure detection and provides clinically meaningful insights into seizure dynamics. The source code is available at \url{https://github.com/LabRAI/IRENE}.

%\url{https://anonymous.4open.science/r/IRENE-C7C8/}
\end{abstract}


\section{Introduction}

Epileptic seizures are a prevalent neurological disease, and their timely detection is essential for clinical diagnosis and intervention~\cite{kalkach2024seizure,loddenkemper2023detect}.
Recent advances in deep learning have significantly improved the automation of EEG analysis for seizure detection~\cite{SNNchen2023,BIOT_NEURIPS2023,MMM_NEURIPS2023,SODorAAAI2025}. In particular, graph neural networks (GNNs) have shown strong capabilities in modeling spatial dependencies and capturing the dynamic topological patterns in multi-channel EEGs~\cite{he2022spatial,li2024dynamical,chen2022brainnet,klepl2024graph}. To leverage this, researchers typically construct graph-based EEG representations, where nodes correspond to EEG channels and edges reflect inter-channel correlations. Through graph learning, these methods capture interactions and detect abnormal patterns among brain regions, thereby improving seizure detection performance and facilitating the discovery of more informative biomarkers.

While graph-based modeling holds great potential, a critical research question lies in how the EEG graph structure is constructed and represented. In existing methods, the graph is typically defined using pairwise similarity or correlation between EEG channels~\cite{ho2023self,klepl2024graph}, estimated from raw signals or extracted features~\cite{tang2022selfsupervised}. This structure is then fixed during training and not subject to optimization. Such modeling implicitly assumes that the initial graph can reflect underlying neural interactions. However, EEG data are notoriously noisy~\cite{ozkahraman2024impact,lim2024baseline}, and no universally effective filtering or feature extraction technique currently exists to guarantee robust graph construction. As a result, the predefined graphs are usually suboptimal, embedding spurious or irrelevant connections. Since GNNs heavily rely on the provided graph to guide message passing and representation learning~\cite{kipf2016semi,wu2019simplifying,dong2023towards}, a poor graph structure inevitably leads to degraded model performance and unreliable seizure detection.

% Several recent efforts have taken initial steps to tackle this issue through graph structure learning. For instance, DCRNN~\cite{tang2022selfsupervised} constructs graphs via normalized cross-correlation between EEG signals. GraphS4mer~\cite{pmlr-v209-tang23a} adopts state space model in combination generic attention mechanisms. Despite these advances, such methods still heavily rely on the quality of the initial graph and lack mechanisms to explicitly refine or correct suboptimal structures during training. Therefore, in this paper, we investigate the novel and critical problem of developing a principled dynamic graph learning framework for EEG seizure detection that can explicitly construct and refine task-relevant, physiologically meaningful brain networks under noisy and label-scarce conditions. This is a non-trivial task as it requires disentangling informative neural interactions from highly noisy EEG signals while ensuring the learned graph structures are compact, interpretable, and generalizable across patients and recording sessions.

Given these challenges with existing EEG graph construction methods, our work aims to systematically address the following three fundamental challenges. \textit{First, how to conduct pysiologically-plausible graph structure learning:} Effective seizure detection from EEG relies on identifying which electrodes (nodes) are involved and how seizure activities propagate through the brain network~\cite{liu2022localization}. However, most existing graph learning-based methods treat the dynamic graphs as latent representations optimized indirectly during training~\cite{ho2023self,ZHAO2021106277}, without explicit mechanisms to enforce physiological plausibility. \textit{Second, how to jointly leverage the learned graph structure and EEG representations for accurate seizure detection?} Existing methods often decouple graph construction from downstream tasks, relying on generic GNNs or attention mechanisms that overlook the reliability or clinical relevance of the connections. This separation limits the ability to suppress noise and the utilization of domain-informed structure.\textit{Third, generalizable seizure detection under label scarcity and data variability:} obtaining high-quality seizure annotations is highly expensive and requires expert clinical knowledge~\cite{wong2025channel,yang2022continental}, leading to a large number of EEG sequences without precise labels (e.g., seizure onset and offset mark)~\cite{raeisi2023class}. Also, EEG signals exhibit high inter-personal variability and are sensitive to differences in acquisition settings~\cite{chen2022brainnet,peng2022seizure}, which pose challenges for model generalization. These challenges highlight the need for a principled framework that constructs task-informative and interpretable graphs from EEG signals, and learns structure-aware spatial-temporal representations through self-supervised training.

To address the aforementioned challenges, we propose IRENE, a novel framework that integrates self-supervised learning with information bottleneck-based dynamic graph modeling to enable interpretable and robust seizure detection. \textit{To address the first challenge}, we explicitly construct task-informative graphs by optimizing an Information Bottleneck (IB) objective, which is different from prior approaches that implicitly learn graph structure. The IB objective encourages sparse yet discriminative edge connections by balancing relevance to labels and compression of input, leading to sparse and interpretable graph representations. \textit{To address the second challenge}, we introduce a graph structure-aware attention mechanism (GSA-Attn) into IRENE's encoder network, which explicitly incorporates graph patterns learned through the IB objective-guided graph construction. Specifically, we integrate edge confidence scores derived from self-expressive coefficient into the attention computation. This allows the model to prioritize physiologically meaningful connections, while suppressing noisy or unstable interactions. \textit{To address the third challenge}, we adopt a self-supervised learning paradigm based on Graph Masked AutoEncoder architecture. IRENE is pre-trained by masking then reconstructing the node attributes of EEG graphs derived from IB principles. This enables IRENE's Encoder to learn generalized and physiologically grounded spatial-temporal representations without relying on labeled seizure data. By leveraging structural priors during pretraining, the model becomes more robust to variations across patients and recording conditions. Our main contributions are three-fold:
\begin{itemize}[leftmargin=15pt,labelsep=5pt]
    \item \textbf{Information Bottleneck-guided dynamic graph construction.} We propose a principled framework that explicitly learns task-relevant and interpretable brain connectivity graphs. By optimizing a mutual information objective that balances label relevance and input compression, our method constructs sparse yet discriminative edge structures, enabling both accurate seizure detection and enhanced clinical interpretability.
    \item \textbf{IRENE -- A Self-Supervised Graph Masked Autoencoder Framework.} IRENE is proposed to tackle label-scarce and inter-patient variability conditions. By reconstructing node features from IB-guided graphs, IRENE captures robust spatial-temporal dependencies while mitigating noise. We further introduce a GSA-Attn mechanism that leverages edge confidence scores, enabling the model to prioritize physiologically meaningful connections during information propagation.
    \item \textbf{Comprehensive Experimental Evaluation.} Extensive experiments on the TUSZ benchmark demonstrate that IRENE consistently outperforms state-of-the-art EEG graph learning methods in seizure detection and classification tasks. Our model achieves superior accuracy and robustness in handling patient variability and adapting to various clinical settings with great interpretability. We confirm the effectiveness of each model component through ablation and robustness studies.
\end{itemize}

\section{Notations}

\noindent \textbf{Graph-Based EEG Representation:} EEG signals naturally exhibit non-Euclidean, graph-like characteristics due to the spatial and functional relationships among EEG channels (i.e., electrodes). Thus, EEG signals can be modeled as a sequence of dynamic graphs. Specifically, at each time step $t$, the brain activity is represented as graph $\mathcal{G}_t = (\mathcal{V}, \mathcal{A}_t, \mathcal{X}_t)$, where $\mathcal{V} = \{\mathbf{v}_1, \mathbf{v}_2, ..., \mathbf{v}_N\}$ is the set of $N$ EEG channels as graph nodes. $\mathcal{A}_t \subseteq \mathcal{V} \times \mathcal{V}$ represents the edge set at time $t$, capturing the dynamic connectivity relationships. Each node $\mathbf{v}_i$ is associated with a feature matrix $\mathbf{x}_t^{i}$, representing the temporal and spatial characteristics of the EEG channel within the current time window. The collection of all node features at time step $t$ is denoted as $\mathcal{X}_t = \{\mathbf{x}_t^1, \mathbf{x}_t^2, ..., \mathbf{x}_t^N\}$.

\noindent \textbf{Information Bottleneck (IB):} The IB principle is an information-theoretic framework for representation learning, which aims to extract a compact yet informative representation from input data by balancing sufficiency and minimality~\cite{tishby2000information,alemi2016deep}. Given an input variable $\mathbf{x}$ and its associated label $\mathbf{y}$, IB seeks to learn a stochastic representation $\mathbf{z}$ that preserves the information relevant for predicting $\mathbf{y}$ while discarding redundant information from $\mathbf{x}$. Formally, the IB objective can be formulated as: $\min_{p(\mathbf{z}|\mathbf{x})} \mathcal{L}_{\mathrm{IB}} = I(\mathbf{x};\mathbf{z}) - \beta I(\mathbf{z};\mathbf{y})$, where $I(\mathbf{x};\mathbf{z})$ is the mutual information between the input $\mathbf{x}$ and latent representation $\mathbf{z}$, encouraging compression of input information. $I(\mathbf{z};\mathbf{y})$ measures the predictive power of $\mathbf{z}$ with respect to $\mathbf{y}$. $\beta>0$ is a trade-off hyperparameter balancing compression and informativeness. $\mathbf{z}$ is a latent representation extracted from the input data $\mathbf{x}$ using a neural network. In graph learning scenario, a popular approach is to rewrite the standard IB equation as: $\min_{p(\mathbf{z}|\mathcal{G}_\mathbf{x})} \mathcal{L}_{\mathrm{IB}} = I(\mathcal{G}_\mathbf{x}; \mathbf{z}) - \beta I(\mathbf{z}; \mathbf{y})$, where $\mathcal{G}_\mathbf{x}$ denotes the graph data, $\mathbf{z}$ denotes the latent representation.

\begin{problem}[Dynamic Graph-based Seizure Detection.]
Given a sequence of dynamic brain graphs $\mathcal{G}_t = (\mathcal{V}, \mathcal{A}_t, \mathcal{X}_t), t=1,2,...,T$, where $\mathcal{V}$ denotes the set of EEG channels as nodes, $\mathcal{A}_t$ and $\mathcal{X}_t$ denote the edge set and node feature set at time step $t$, respectively. Our goal is to predict the corresponding seizure label $y_t$ for each time step $t$ by designing a function $f$ that maps the input graph sequence to the label sequence, i.e., $f: \{\mathcal{G}t\}_{t=1}^T \rightarrow \{y_t\}_{t=1}^T$, through minimizing the seizure detection or classification error over the temporal sequence.
\end{problem}

\section{Methodology}

%In this section, we present the full methodology of our proposed IRENE framework, which integrates dynamic graph construction, structure-aware encoding, and self-supervised representation learning in a unified architecture. The detailed illustration of IRENE framework is provided in Figure~\ref{fig1-EEG-description}, it comprises three core components: (1) an Information Bottleneck-guided dynamic graph constructor that explicitly learns sparse and task-informative brain connectivity graphs from raw EEG signals; (2) a structure-aware encoder network equipped with a specialized attention mechanism (GSA-Attn), which incorporates edge confidence scores into the message passing process to emphasize reliable, physiologically meaningful dependencies; and (3) a Graph Masked AutoEncoder pretraining scheme that reconstructs masked node features based on the learned graphs, guided by a novel loss function that jointly preserves attribute fidelity and structural consistency.

\begin{figure*}[t]
  \centering
  \includegraphics[width=0.9\textwidth]{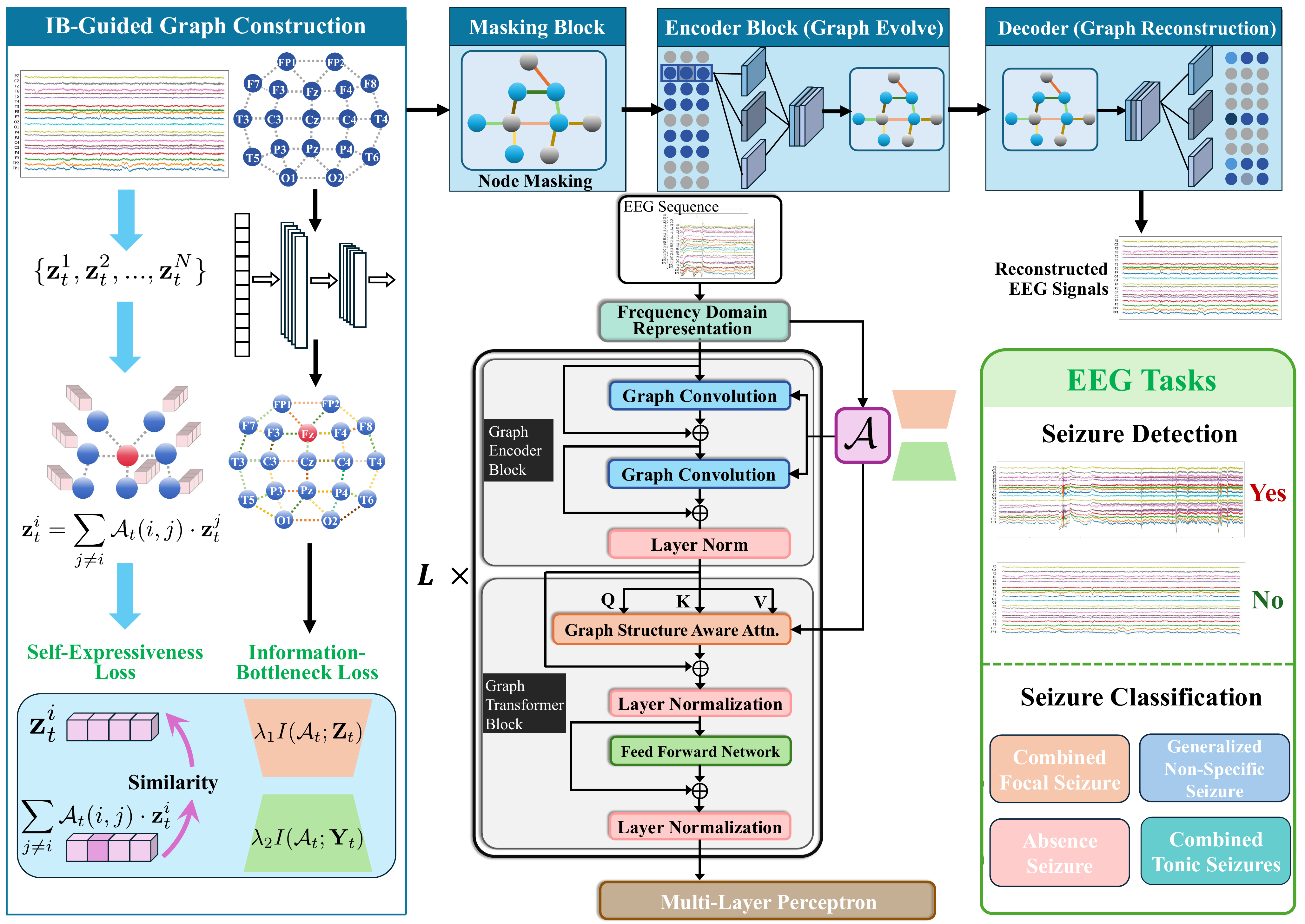}
  \caption{The architecture of the proposed IRENE framework.}
  \label{fig1-EEG-description}
\end{figure*}

\subsection{Model Architecture}
\textbf{IRENE Model Architecture.} Figure~\ref{fig1-EEG-description} presents an overview of our IRENE framework, which integrates IB-guided dynamic graph construction with a structure-aware encoder-decoder architecture. The entire pipeline begins with raw EEG signals, which are first transformed into the frequency domain to better capture meaningful neural oscillations. These processed signals are then passed into our \textit{IB-Guided Graph Construction} module (left block in the figure), where an initial set of node embeddings $\{z_t^1, z_t^2, \dots, z_t^N\}$ is extracted for each time step $t$ using a shared encoder. These embeddings are used to learn a dynamic adjacency matrix $\mathcal{A}_t$ via a self-expressiveness formulation regularized by an information bottleneck (IB) loss. The loss encourages the graph to maintain minimal yet task-relevant dependencies among EEG channels. Once the dynamic graph structure is obtained, it is input into the encoder-decoder backbone of IRENE for representation learning. The center block of the figure shows the graph encoder composed of two major components: (i) a \textit{Graph Encoder Block}, where local spatial interactions are captured using stacked graph convolutional layers guided by $\mathcal{A}_t$; and (ii) a \textit{Graph Transformer Block}, which leverages structure-aware self-attention to model global brain interactions. The resulting node-level representations are then processed through a Multi-Layer Perceptron (MLP) for downstream seizure-related tasks. Notably, the entire model is trained end-to-end with both reconstruction and supervision signals.

\textbf{Graph Learning and Model Workflow.} The core learning procedure of IRENE follows a masked graph autoencoder paradigm designed to exploit the dynamics and topology of brain connectivity. As shown in the top pathway of the architecture, given an input EEG clip, we first construct a dynamic graph at each time $t$ using our IB-guided strategy. This involves jointly minimizing a self-expressiveness loss $||\mathbf{Z}_t - \mathbf{Z}_t\mathcal{A}_t||_F^2$ to capture relational dependencies, while simultaneously optimizing two mutual information terms $\lambda_1 I(\mathcal{A}_t;\mathbf{Z}_t)$ for compressing redundant connections and $\lambda_2 I(\mathbf{A}_t; Y_t)$ for maximizing task relevance. The learned graph $\mathcal{A}_t$ is then passed into the encoder, which consists of $L$ stacked blocks. In each block, the Graph Encoder Block performs local graph convolution guided by $\mathcal{A}_t$, while the Graph Transformer Block applies a structure-aware attention mechanism where edge weights modulate attention scores across channels. During self-supervised pretraining, a masking block randomly occludes a subset of node features, and the decoder aims to reconstruct the original features based on the learned context. This denoising-style objective enhances the model's ability to extract robust, informative EEG representations. The final embeddings can be fine-tuned for diverse EEG analysis tasks (right block), including seizure detection and multi-class seizure classification, demonstrating IRENE’s generalization and interpretability.

\subsection{Information Bottleneck-Guided Dynamic Graph Construction}\label{sec-3-1}

EEG signals are inherently noisy and exhibit time-varying inter-channel interactions~\cite{fan2024eeg,li2022graph}. Motivated by the Information Bottleneck (IB) theory~\cite{sun2022graph,wei2022contrastive}, we propose a principled dynamic graph construction method that extracts denoised and task-informative graph structures. At each time step $t$, we encode the feature vector of each EEG channel into a latent representation $\mathbf{z}_t^i \in \mathbb{R}^{d \times T}$ using a shared encoder $f_{\boldsymbol{\theta}}$. Collectively, these node-wise embeddings form a brain state collection $\mathbf{Z}_t = \{\mathbf{z}_t^1, \mathbf{z}_t^2, \dots, \mathbf{z}_t^N\} \in \mathbb{R}^{N \times d \times T}$, where each $\mathbf{z}_t^i$ is treated as a distinct view of the whole underlying brain state at this moment.

\textbf{Self-Expressive Graph Construction via Information Bottleneck.}
To uncover inter-channel dependencies while suppressing noise, we adopt a self-expressiveness formulation guided by an IB objective. Each node embedding is approximated as a sparse, weighted combination of other nodes, where the magnitude of each weight reflects the strength of the corresponding edge connection:
\begin{equation}
\mathbf{z}_t^i = \sum_{j \neq i} \mathbf{A}_t(i,j) \cdot \mathbf{z}_t^j
\label{eq:self-express}
\end{equation}

\noindent Here, $\mathbf{A}_t \in \mathbb{R}^{N \times N}$ is the learnable graph adjacency matrix at time $t$, which encodes the inter-channel connectivity. Each coefficient $\mathbf{A}_t(i,j)$ reflects the importance of node $j$ in reconstructing node $i$, and its magnitude $\phi_{ij} = |\mathbf{A}_t(i,j)|$ is used as the structure confidence score for the attention mechanism gating, which we will introduce in section~\ref{model_architecture}.

\textbf{Information Bottleneck Loss for Structure Learning.}
To ensure the learned graph captures compact yet discriminative structural dependencies, we treat $\mathbf{A}_t$ as the bottleneck variable and formulate an IB-based objective:
{\footnotesize
\begin{equation}
\mathcal{L}_{\text{IB-Graph}} = \underbrace{\|\mathbf{Z}_t - \mathbf{Z}_t \mathbf{A}_t\|_F^2}_{\text{Self-expressiveness loss}} + \underbrace{\lambda_1 I(\mathbf{A}_t; \mathbf{Z}_t)}_{\text{Redundancy regularization}} - \underbrace{\lambda_2 I(\mathbf{A}_t; \mathbf{Y}_t)}_{\text{Information maximization}}
\label{eq:ib-graph}
\end{equation}
}
\noindent Since our goal is to optimize only the graph structure, we explicitly treat $\mathbf{A}_t$ as the information bottleneck variable. The first term encourages graph-based reconstruction of node features. The second term penalizes redundant structure by discouraging mutual information between $\mathbf{A}_t$ and $\mathbf{Z}_t$. The third term maximizes the predictive power of $\mathbf{A}_t$ for the seizure label $\mathbf{Y}_t$. Hyperparameters $\lambda_1$ and $\lambda_2$ balance compression and discriminativeness.

\textbf{Temporal Consistency Regularization.} 
Given the inherent temporal continuity of neural activities, brain functional connectivity is expected to evolve smoothly over time rather than exhibiting abrupt structural changes~\cite{thompson2017brain}. To align the learned dynamic graphs with this neurophysiological prior and encourage the smooth evolution of brain connectivity, we introduce a temporal consistency regularization that encourages gradual transitions in graph topology.~\cite{kong2024learning,bao2024dynamic}
Specifically, we regularize the change of graph topology in adjacent time steps, resulting in the temporal smoothness regularization: $\mathcal{L}_{\text{smooth}}=\|\mathbf{A}_t - \mathbf{A}_{t-1}\|_F^2$. This equation is based on Frobenius norm, which penalizes abrupt changes in the connectivity matrices between consecutive time steps~\cite{liu2016learning}.

\textbf{Final Dynamic Graph Construction.}  
After training, we retain the learned adjacency matrix $\mathbf{A}_t$ as a soft graph with continuous edge weights. To ensure sparsity while preserving the relative importance of connections, we first apply Min-Max normalization to $|\mathbf{A}_t(i,j)|$, which maps values to the range [0,1]. Subsequently, for each node, we apply a Top-K sparsification strategy. The resulting sparse yet weighted \textbf{final adjacency matrix:} $\tilde{\mathbf{A}}_t \in [0,1]^{N \times N}$. The refined graph $\mathcal{G}_t = (\mathcal{V}, \tilde{\mathbf{A}}_t, \mathbf{Z}_t)$ serves as the structural input to IRENE's encoder. Furthermore, the edge weights $\tilde{\mathbf{A}}_t(i,j)$ are used as structure-aware confidence scores $\phi_{ij}$ to softly gate the proposed GSA-Attn mechanism. 
%\yd{DO NOT HAVE ANY dangling word throughout this paper.}

To make the Information Bottleneck objective in Eq.~(\ref{eq:ib-graph}) computationally tractable, we employ variational estimation techniques to approximate the mutual information terms $I(\mathbf{A}_t;\mathbf{Y}_t)$ and $I(\mathbf{A}_t;\mathbf{Z}_t)$. Specifically, we adopt a contrastive lower bound to estimate the predictive term $I(\mathbf{A}_t; \mathbf{Y}_t)$, and an adversarial upper bound to estimate the redundancy term $I(\mathbf{A}_t;\mathbf{Z}_t)$.

\noindent\textbf{Estimating $I(\mathbf{A}_t;\mathbf{Y}_t)$:} We approximate this mutual information using InfoNCE bound~\cite{tschannen2019mutual}, which transforms the estimation into a contrastive learning task. Let $g_\phi(\mathbf{A}_t,\mathbf{Y}_t)$ denote a learned scoring function (e.g., a bilinear classifier), then the lower bound is:
\begin{equation}
I(\mathbf{A}_t;\mathbf{Y}_t) \geq \mathbb{E}\left[ \log \frac{\exp(g_\phi(\mathbf{A}_t,\mathbf{Y}_t))}{\sum_{\mathbf{y}' \in \mathcal{N}} \exp(g_\phi(\mathbf{A}_t,\mathbf{y}'))} \right]
\label{eq:info-predict}
\end{equation}
where $\mathcal{N}$ is a set of negative samples drawn from the label distribution. This encourages the graph structure to retain seizure-discriminative information.

\textbf{Estimating $I(\mathbf{A}_t;\mathbf{Z}_t)$:} We estimate it using a variational upper bound based on the Donsker-Varadhan (DV) representation of the KL divergence~\cite{ghimire2021reliable}. Let $T_\psi(\mathbf{A}_t,\mathbf{Z}_t)$ be a critic function parameterized by a neural network. The upper bound becomes:
\begin{equation}
I(\mathbf{A}_t;\mathbf{Z}_t) \leq \mathbb{E}_{P(\mathbf{A}_t,\mathbf{Z}_t)}[T_\psi] - \log \mathbb{E}_{P(\mathbf{A}_t)P(\mathbf{Z}_t)}[\exp(T_\psi)]
\label{eq:info-redundancy}
\end{equation}
This penalizes unnecessary dependency between $\mathbf{A}_t$ and potentially noisy or irrelevant input features.

\subsection{Structure-Aware Graph Transformer for EEG Representation Learning}\label{model_architecture}

We implement IRENE as a structure-aware Graph Masked AutoEncoder, designed to model seizure-discriminative dependencies from IB-guided dynamic graphs in a self-supervised fashion. Its encoder adopts a Transformer-style backbone that stacks $L$ layers of residual blocks, each composed of two submodules: a graph convolution module and a structure-aware soft mask attention mechanism.

\textbf{Masking Strategy.} During pretraining, we adopt a node-wise masking strategy tailored to EEG signals. At each time step $t$, we randomly select a subset $\mathcal{M} \subset \mathcal{V}$ of EEG channels to be masked. For each masked node ${v}_i \in \mathcal{M}$, its input feature $\mathbf{x}_i$ is replaced by a fixed token—either a zero vector or a learned embedding—to prevent direct leakage of information. Each node’s feature corresponds to a temporal patch from its EEG signal, making the masking operation patch-level at the channel scale. The model is trained to reconstruct the original node features $\{\mathbf{x}_i \in \mathcal{M}\}$ from the remaining unmasked nodes and the current graph structure $\mathcal{G}_t$, thereby enforcing structure-aware predictive learning. Unless otherwise stated, the masking ratio is fixed at 15\% throughout pretraining.

\textbf{Encoder Architecture.} At each time step $t$, the encoder receives the dynamic graph $\mathcal{G}_t = (\mathcal{V}, \mathbf{A}^{*}_t,\mathbf{Z}_t)$, where $\mathbf{Z}_t \in \mathbb{R}^{N \times d \times T}$ is the node embedding matrix and $\mathbf{A}^{*}_t$ is the top-$k$ sparsified binary adjacency matrix derived from the IB-guided graph constructor. Each encoder block begins with two stacked GCN layers to capture local neighborhood information under the structural prior $\mathbf{A}^{*}_t$. The resulting representation is then passed to a full-graph self-attention module, which enables long-range interaction across EEG channels. Residual connection and layer normalization are applied after each submodule to enhance stability and gradient flow, following standard Transformer design.

\textbf{Graph Structure-Aware Attention.} To incorporate the IB-derived graph structure into attention computation, we introduce a graph structure-aware (GSA) attention mechanism. For each node pair $(i,j)$, the attention score is computed as:
\begin{equation}
\alpha_{ij} = \text{softmax}_j \left( \frac{\mathbf{q}_i^\top \mathbf{k}_j}{\sqrt{d}} + \gamma \cdot \phi_{ij} \right)
\label{eq:soft-mask-att}
\end{equation}
Here, $\mathbf{q}_i$ and $\mathbf{k}_j$ denote the query and key vectors of node $i$ and $j$, respectively, computed via linear projections of the input node representations. $\gamma$ is a learnable parameter that balances the influence of the structural prior. $\phi_{ij}$ is the structure confidence score derived as $\phi_{ij} = |\mathbf{A}_t(i,j)|$, where $\mathbf{A}_t(i,j)$ is the self-expressiveness coefficient learned from the IB-guided graph constructor.

Once attention weights $\alpha_{ij}$ are obtained, each node updates its representation by aggregating information from its neighbors, weighted by attention scores: $\mathbf{h}_i' = \sum_{j=1}^{N} \alpha_{ij} \cdot \mathbf{h}_j$, where $h_j$ is the value vector of node $j$, also obtained via a learned linear transformation of the input embedding $h_j$. The updated feature $h_i'$ captures both content-based relevance and structural plausibility (via $\phi_{ij}$). The attention mechanism is followed by a residual connection and layer normalization: $\tilde{\mathbf{h}}_i = \text{LayerNorm}(\mathbf{h}_i + \mathbf{h}_i')$. This formulation allows IRENE to softly constrain its attention flow using biologically grounded structural priors, while retaining the flexibility of full-graph attention to discover non-obvious dependencies. As a result, the encoder learns robust and physiologically meaningful EEG representations under both noisy and label-scarce conditions.

\subsection{Model Training Strategy and Loss Function}

We adopt a two-stage training strategy that separates representation pretraining from downstream seizure classification. This approach enables the encoder to first learn robust, structure-aware EEG features in a self-supervised manner before adapting to the final prediction task.

% \textbf{Stage 1: Pretraining with Self-Supervised Objectives.}  
% In the first stage, we train IRENE's encoder using both the reconstruction objective and the Information Bottleneck-guided graph construction losses. The total loss for pretraining is given by:

% \begin{equation}
% \mathcal{L}_{\text{pretrain}} = \mathcal{L}_{\text{IB-Graph}} + \lambda_3 \mathcal{L}_{\text{smooth}} + \lambda_4 \mathcal{L}_{\text{recon}}
% \label{eq:pretrain-loss}
% \end{equation}

% Here, $\mathcal{L}_{\text{IB-Graph}}$ enforces compactness and task-relevance of the learned dynamic graph structures, $\mathcal{L}_{\text{smooth}}$ promotes temporal consistency of connectivity, and $\mathcal{L}_{\text{recon}}$ trains the encoder to reconstruct masked EEG patches from the observed graph context \yd{Objective function does not train things. the subjective does not match the verb in this sentence. REVISE ALL SIMILAR PLACES THROUGHOUT THIS PAPER. Also, introduce what is used to instantiate $\mathcal{L}_{\text{recon}}$, MSE loss?}.

\noindent \textbf{Stage 1: Pretraining with Self-Supervised Objectives.}  
In the first stage, we optimize IRENE's encoder by jointly minimizing a reconstruction objective and the Information Bottleneck-guided graph construction losses. The total pretraining loss is defined as:
\vspace{-1mm}
\begin{equation}
\mathcal{L}_{\text{pretrain}} = \mathcal{L}_{\text{IB-Graph}} + \lambda_3 \mathcal{L}_{\text{smooth}} + \lambda_4 \mathcal{L}_{\text{recon}}
\label{eq:pretrain-loss}
\end{equation}

Here, $\mathcal{L}_{\text{IB-Graph}}$ enforces the compactness and task-relevance of the learned dynamic graph structures, while $\mathcal{L}_{\text{smooth}}$ promotes temporal consistency by regularizing abrupt topology changes across adjacent time steps. The reconstruction loss $\mathcal{L}_{\text{recon}}$ encourages the encoder to learn structure-aware representations that can predict missing node features from the surrounding graph context. Specifically, given a randomly masked subset of EEG channels $\mathcal{M} \subset \mathcal{V}$ at each time step, we reconstruct the original node features $\{\mathbf{x_i}\}_{i \in \mathcal{M}}$ from their latent embeddings using a shared MLP $\text{Decoder}_{\psi}(\cdot)$:
\begin{equation}
\mathbf{\hat{x}_i} = \text{Decoder}_{\psi}(\mathbf{h_i}), \quad \forall i \in \mathcal{M}
\end{equation}

The reconstruction loss is computed as the Mean Squared Error (MSE) between the reconstructed and original features of the masked nodes: $\mathcal{L}_{\text{recon}} = \frac{1}{|\mathcal{M}|} \sum_{i \in \mathcal{M}} \| \mathbf{\hat{x}_i} - \mathbf{x_i} \|_2^2$. This objective forces the encoder to capture discriminative and robust spatial dependencies, enhancing its ability to generalize under noisy and label-scarce conditions.

\noindent \textbf{Stage 2: Fine-Tuning for Seizure Classification.}  
Once pretraining is complete, we fix the encoder parameters, and attach a task-specific classification head (e.g., a multi-layer perceptron) to the encoder. The model is then fine-tuned using the Cross-Entropy Loss over true labels. This two-stage setup allows the model to first develop generalized graph-aware representations and then specialize to seizure classification, enhancing both robustness and discriminative power.

\section{Experimental Evaluation}\label{sec:experiment}

In this section, we conduct extensive experiments to evaluate the effectiveness, robustness, and interpretability of our proposed method. Specifically, we aim to address the following research questions: \textbf{RQ1.} {Does our method show improve performance compared to state-of-the-art EEG graph learning baselines?} \textbf{RQ2.} {To what extent does the IB principle reduce redundant edges in the constructed graphs, and how does this affect model interpretability?} \textbf{RQ3.} {How do different components of IRENE contribute to its overall performance?} In the following sections, we first introduce the experimental settings and datasets, followed by detailed results and analysis to answer each of the above questions.

\subsection{Experimental Settings} \label{sec:exp-setting}
\noindent \textbf{Dataset.} We conduct experiments on the publicly available EEG seizure benchmark dataset: the Temple University Hospital EEG Seizure Corpus (TUSZ) v1.5.2~\cite{shah2018temple, rahman2020improving}. TUSZ is currently one of the largest clinical EEG corpora for seizure detection, comprising 5,612 EEG recordings and 3,050 seizure annotations. It includes 19 EEG channels recorded using the standard 10-20 system, covering a broad range of subjects across diverse clinical conditions. For additional interpretability analyses, we leverage the seizure onset-offset labels and event type annotations provided in TUSZ dataset.

\noindent \textbf{Baseline Methods.} To comprehensively evaluate the effectiveness of our proposed method, we compare it against a diverse set of baseline models spanning four categories. (i) As classic deep learning baselines without graph modeling, we include the standard LSTM network~\cite{hochreiter1997long} and ResNet-LSTM~\cite{lee2022real}, which combines residual convolutional layers with temporal recurrence for seizure detection. (ii) For static graph-based modeling, we evaluate Dist-DCRNN~\cite{tang2022selfsupervised}, which builds a distance-based fixed connectivity graph and applies diffusion convolutional recurrent network. (iii) To benchmark against dynamic graph learning approaches tailored for EEG, we include GRAPHS4MER~\cite{pmlr-v209-tang23a}, NeuroGNN~\cite{hajisafi2024dynamic}, and Corr-DCRNN~\cite{tang2022selfsupervised}, all of which learn time-evolving brain connectivity graphs to capture transient neural dynamics.

\begin{table*}[!htbp]
\centering
\footnotesize
\renewcommand{\arraystretch}{1.0}
\caption{Performance comparison among different methods on seizure detection task.}
\resizebox{0.95\linewidth}{!}{
\begin{tabular}{l|ccc|ccc}
\toprule
\multirow{2}{*}{\textbf{Method}} & \multicolumn{3}{c|}{\textbf{Detection-12s}} & \multicolumn{3}{c}{\textbf{Detection-60s}} \\
\cmidrule(r){2-4} \cmidrule(l){5-7}
& \textbf{F1} & \textbf{Recall} & \textbf{AUROC} & \textbf{F1} & \textbf{Recall} & \textbf{AUROC} \\
\midrule
LSTM & $0.580 \pm 0.043$ & $0.624 \pm 0.033$ & $0.836 \pm 0.022$ & $0.592 \pm 0.033$ & $0.638 \pm 0.017$ & $0.841 \pm 0.026$ \\
ResNet-LSTM & $0.598 \pm 0.035$ & $0.653 \pm 0.031$ & $0.843 \pm 0.022$ & $0.625 \pm 0.027$ & $0.663 \pm 0.030$ & $0.850 \pm 0.014$ \\
Dist-DCRNN & $0.713 \pm 0.044$ & $0.735 \pm 0.043$ & $0.866 \pm 0.016$ & $0.695 \pm 0.028$ & $0.733 \pm 0.014$ & $0.875 \pm 0.015$ \\
Corr-DCRNN & $0.729 \pm 0.038$ & $0.756 \pm 0.041$ & $0.861 \pm 0.005$ & $0.722 \pm 0.038$ & $0.732 \pm 0.021$ & $0.873 \pm 0.012$ \\
NeuroGNN & $0.647 \pm 0.040$ & $0.710 \pm 0.024$ & $0.865 \pm 0.019$ & $0.698 \pm 0.044$ & $0.733 \pm 0.042$ & $0.871 \pm 0.021$ \\
GraphS4mer & $0.690 \pm 0.034$ & $0.721 \pm 0.025$ & $0.882 \pm 0.014$ & $0.680 \pm 0.012$ & $0.718 \pm 0.041$ & $0.885 \pm 0.012$ \\
\textbf{IRENE (Ours)} & $\mathbf{0.749 \pm 0.032}$ & $\mathbf{0.782 \pm 0.026}$ & $\mathbf{0.908 \pm 0.011}$ & $\mathbf{0.753 \pm 0.028}$ & $\mathbf{0.788 \pm 0.022}$ & $\mathbf{0.916 \pm 0.013}$ \\
\bottomrule
\end{tabular}
}
\label{tab:model_comparison}
\end{table*}

\noindent \textbf{Evaluation Metrics.} To provide a comprehensive and robust evaluation, we adopt three widely-used evaluation metrics in EEG seizure detection and classification~\cite{10196475,chen2022brainnet,tang2022selfsupervised}: Accuracy~\cite{10196475}, F1 score~\cite{chen2022brainnet}, and Area Under the Receiver Operating Characteristic curve (AUROC)~\cite{tang2022selfsupervised}. (i) \textbf{Accuracy} measures the overall proportion of correctly predicted samples out of all predictions. It is the most commonly adopted criteria in classification tasks. (ii) \textbf{F1 score} is the harmonic mean of Precision and Recall, balancing the trade-off between false positives and false negatives. It is particularly suitable for imbalanced EEG datasets. F1 offers a single summary measure that reflects both the model’s precision and sensitivity to seizure events. (iii) \textbf{AUROC} evaluates the model's ability to distinguish between seizure and non-seizure classes across all possible decision thresholds. It is computed as the area under the ROC curve, which plots the true positive rate (recall) against the false positive rate.

\noindent \textbf{Experimental Settings.} All experiments are conducted using an NVIDIA A100 GPU. Adam optimization~\cite{kingma2014adam} with an initial learning rate of $3e^{-4}$ and a weight decay of $1e^{-5}$ is employed in the experiments. Batch size is set to 32 for training and 64 for evaluation. Our model is trained using EEG recordings from TUSZ dataset, which are preprocessed into 12s/60s non-overlapping clips and resampled to a fixed frequency. Input features were z-score normalized using statistics computed on the training set. During training, we balanced seizure and non-seizure samples by random undersampling of the majority class. The model parameters are randomly initialized and optimized to maximize AUROC on the validation set. To ensure robustness, we conduct 5 independent training runs with different random seeds and report the mean performance. Early stopping is applied based on the validation AUROC. For all graph-based models, adjacency matrices are dynamically calculated from EEG signal correlations or loaded from precomputed spatial priors.

\vskip -5pt
\begin{figure*}[t]
    \centering
    % 左侧大图
    \begin{subfigure}[t]{0.45\linewidth}
        \vspace{0pt} % 修正对齐
        \centering
        \includegraphics[height=6.0cm]{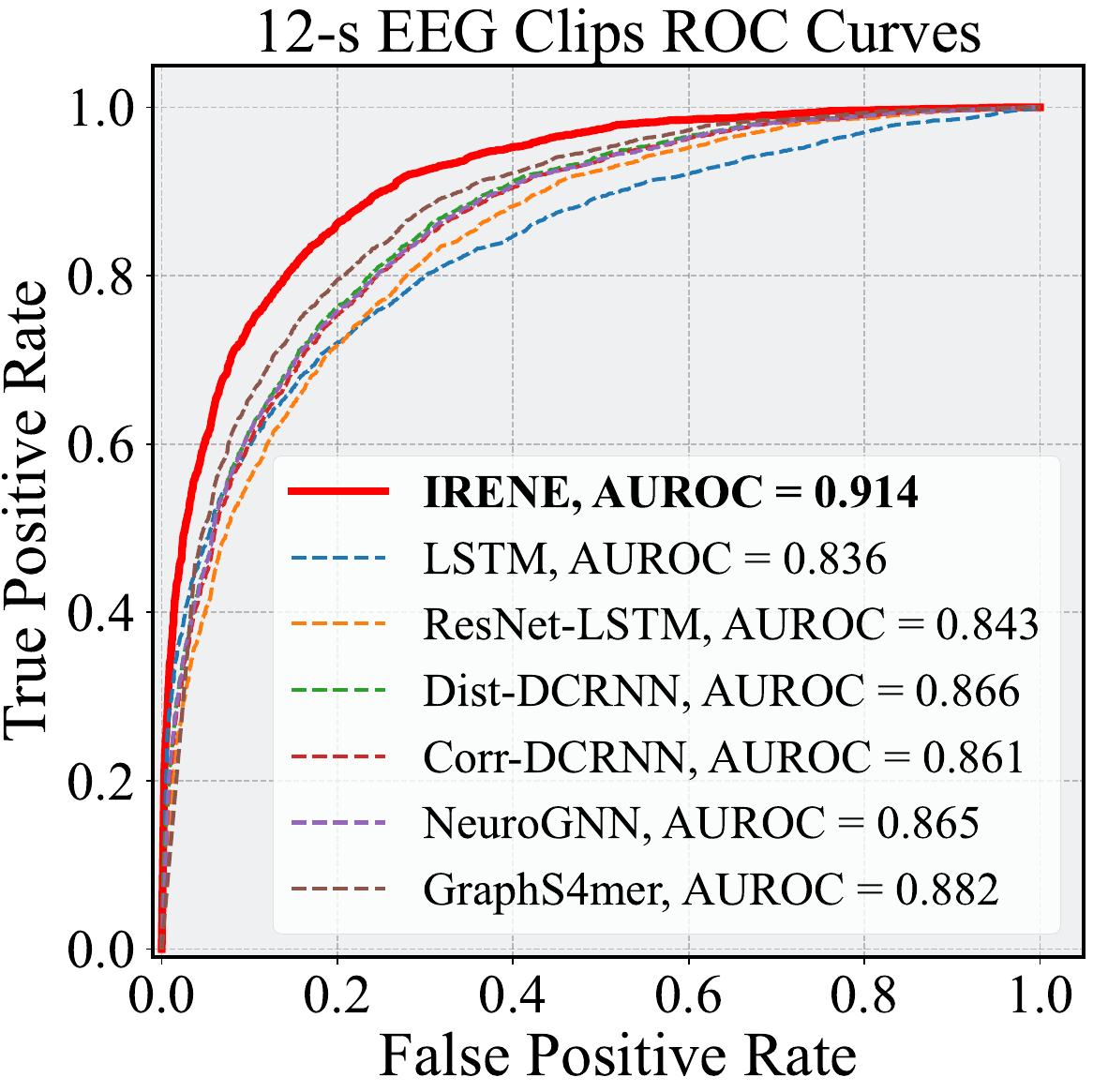}
        \caption{ROC curves}
    \end{subfigure}%
    \hfill
    % 右侧6个小图 (高度与左侧等高)
    \begin{subfigure}[t]{0.55\linewidth}
        \vspace{0pt} % 修正对齐
        \centering
        % 第一行
        \begin{minipage}[t]{0.33\linewidth}
            \centering
            \includegraphics[height=3.0cm]{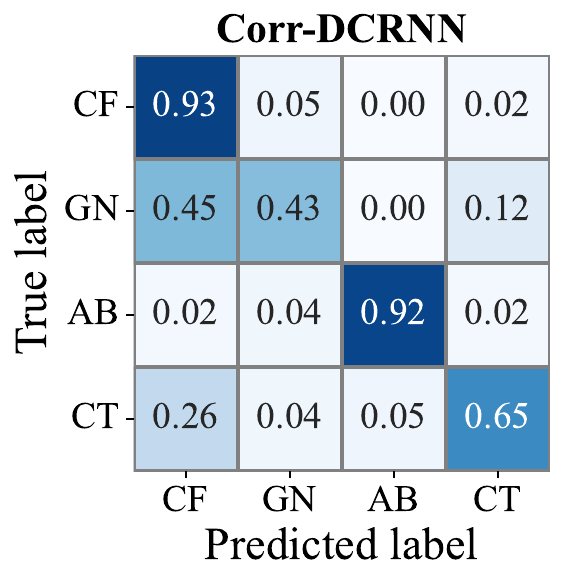}
        \end{minipage}%
        \begin{minipage}[t]{0.33\linewidth}
            \centering
            \includegraphics[height=3.0cm]{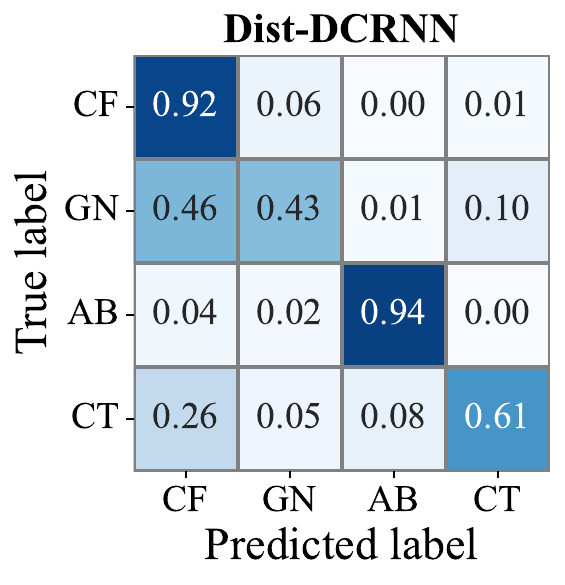}
        \end{minipage}%
        \begin{minipage}[t]{0.33\linewidth}
            \centering
            \includegraphics[height=2.8cm]{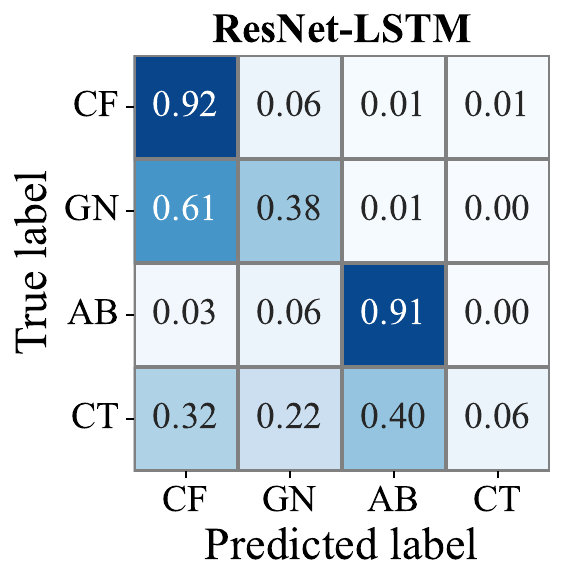}
        \end{minipage}
        % 第二行
        \vspace{1mm}
        \begin{minipage}[t]{0.33\linewidth}
            \centering
            \includegraphics[height=3.0cm]{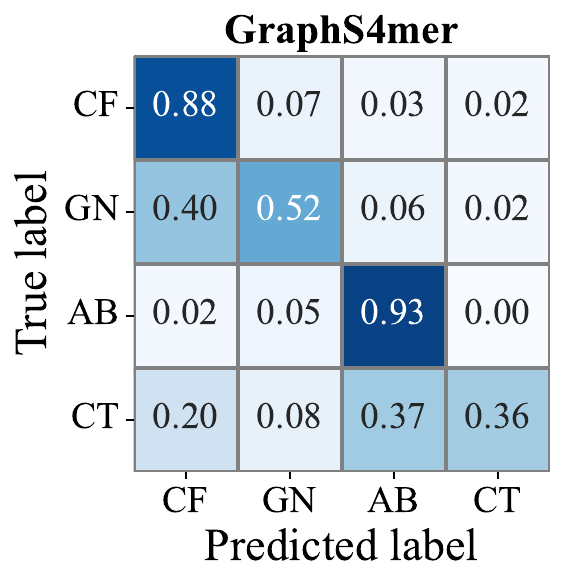}
        \end{minipage}%
        \begin{minipage}[t]{0.33\linewidth}
            \centering
            \includegraphics[height=2.8cm]{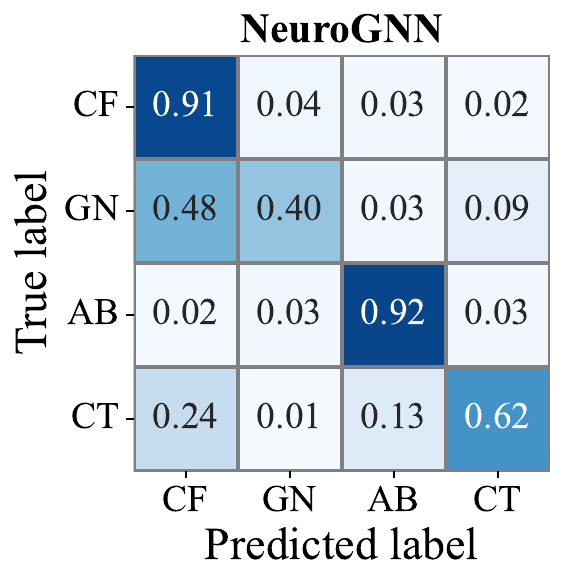}
        \end{minipage}%
        \begin{minipage}[t]{0.33\linewidth}
            \centering
            \includegraphics[height=2.8cm]{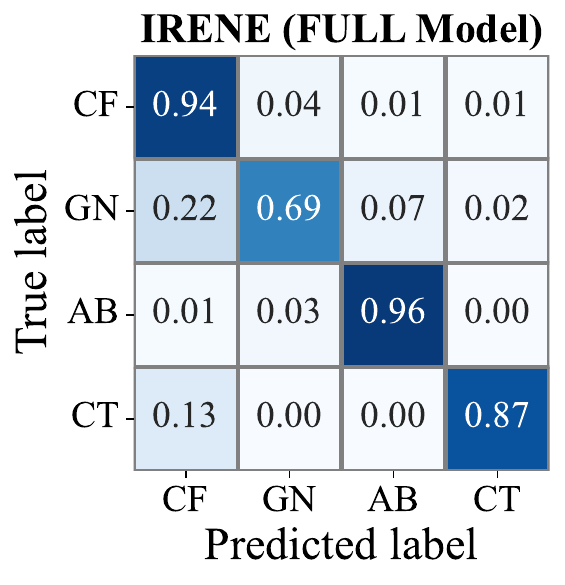}
        \end{minipage}
        \caption{Confusion matrices}
    \end{subfigure}
    \vspace{-2mm}
\caption{(a) ROC curves for seizure detection on 12s EEG clips. (b) Confusion matrices for each baseline model on 12s clip seizure classification. IRENE shows improved per-class prediction accuracy, particularly for the most challenging CT class.}
% \vspace{-2mm}
\label{fig3-exp-ROC}
\end{figure*}

\subsection{Performance Evaluation}

To answer \textbf{RQ1}, we compare IRENE against a suite of representative baselines on seizure detection tasks using the TUSZ dataset, under both 12-second and 60-second EEG clip settings. F1-Score, Recall, and AUROC are used as evaluation metrics, and the results are summarized in Table~\ref{tab:model_comparison}. The ROC curves and confusion matrices are further provided in Figure~\ref{fig3-exp-ROC}(a)-(b). We have the following key observations: (1) IRENE consistently outperforms all baselines across all metrics and clip lengths, achieving the highest F1-Score and AUROC, highlighting its robustness and effectiveness. (2) Traditional RNN-based methods (LSTM, ResNet-LSTM) perform poorly, while graph-based models like Dist-DCRNN and Corr-DCRNN offer notable gains by modeling spatial dependencies, yet still fall short of IRENE. (3) Although GraphS4mer leverages a Transformer backbone, its sensitivity to noise and lack of dynamic graph modeling limit its recall. (4) As shown in Figure~\ref{fig3-exp-ROC}(b), IRENE achieves particularly strong performance in detecting challenging seizure types such as \underline{CF} (Focal Seizure with Consciousness Impairment) and \underline{GN} (Generalized Non-Motor Seizure), while also improving detection of \underline{AB} (Absence Seizure) and \underline{CT} (Clonic Seizure Type). These seizure types exhibit heterogeneous patterns and subtle EEG manifestations, where conventional methods struggle. IRENE demonstrates higher true positive rates and improved accuracy, reflecting its sensitivity to subtle seizure dynamics.

\subsection{Effectiveness of Graph Structure}

%\yd{for all wrapped tables and figures below, only place them into the section where you are discussing them! Do not put anything into the conclusion section.}

%To ANSWER RQ2: To what extent does the IB principle reduce task-irrelevant noise and redundant edges in the constructed graphs, and how does this affect interpretability and discriminative power?
\begin{table}[!htbp]
    \centering
    %\footnotesize
    \vspace{-12pt}
    \caption{Edge density of different graph construction methods integrated with IRENE.}
    \label{exp-RQ2-edge-density}
    \resizebox{0.94\linewidth}{!}{
    \begin{tabular}{>{\raggedright\arraybackslash}p{5.0cm}|>{\centering\arraybackslash}p{2.0cm}}
    \toprule
    \multicolumn{1}{c}{\textbf{Graph Construction Method}} & \multicolumn{1}{c}{\textbf{Edge Density}} \\
    \midrule
    Distance-based $\mathcal{G}$~\cite{tang2022selfsupervised} + IRENE & 0.31 \\
    Cross-Correlation $\mathcal{G}$~\cite{tang2022selfsupervised} + IRENE & 0.72 \\
    Temporal Sim. $\mathcal{G}$~\cite{hajisafi2024dynamic} + IRENE & 0.47 \\
    Semantic Sim. $\mathcal{G}$~\cite{hajisafi2024dynamic} + IRENE & 0.38 \\
    \hline
    \textbf{Original IRERE} (w IB-based $\mathcal{G}$) & \textbf{0.16} \\
    \bottomrule
    \end{tabular}
    }
\end{table}

\begin{figure}[!htbp]
    \centering
    % 左侧大图
    \begin{subfigure}[t]{0.49\linewidth}
        \vspace{0pt} % 修正对齐
        \centering
        \includegraphics[width=0.99\columnwidth]{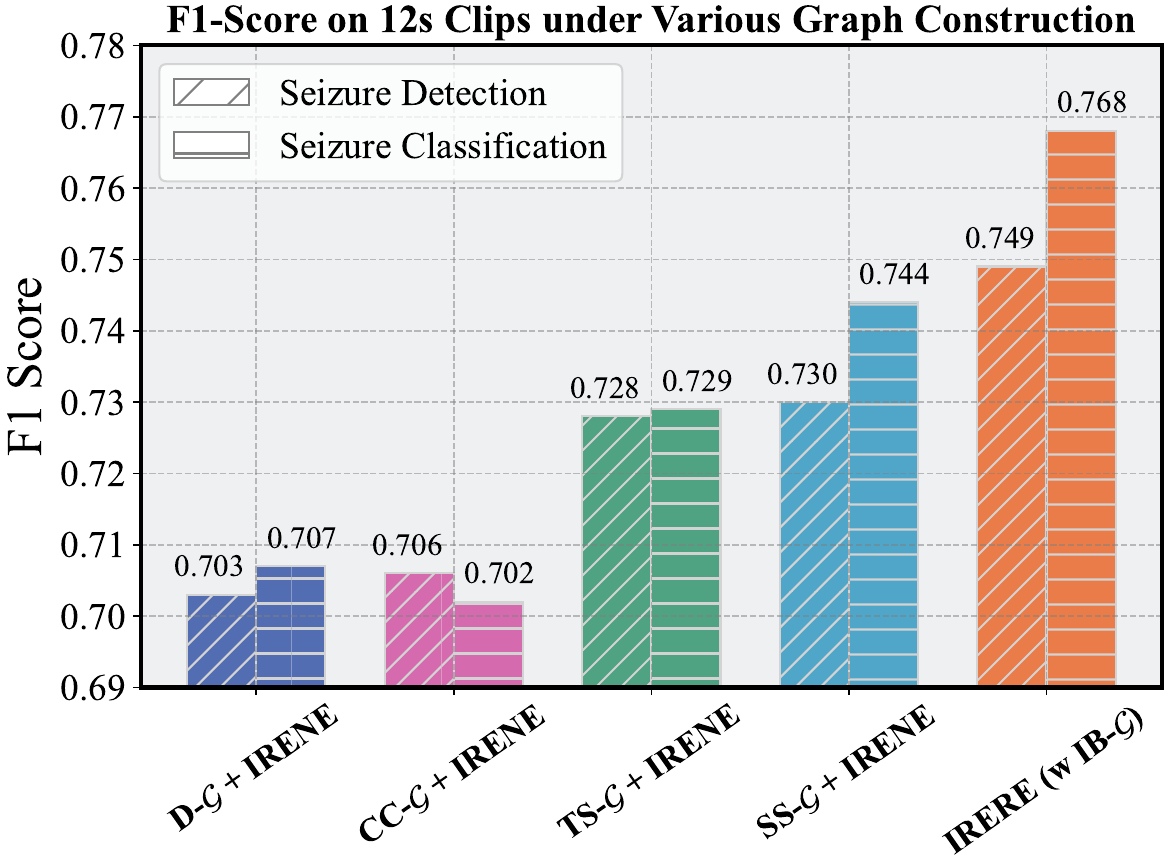}
        \caption{}
        %\caption{F1 across graph construction methods (12s)}
    \end{subfigure}%
    \begin{subfigure}[t]{0.49\linewidth}
        \vspace{0pt} % 修正对齐
        \centering
        \includegraphics[width=0.99\columnwidth]{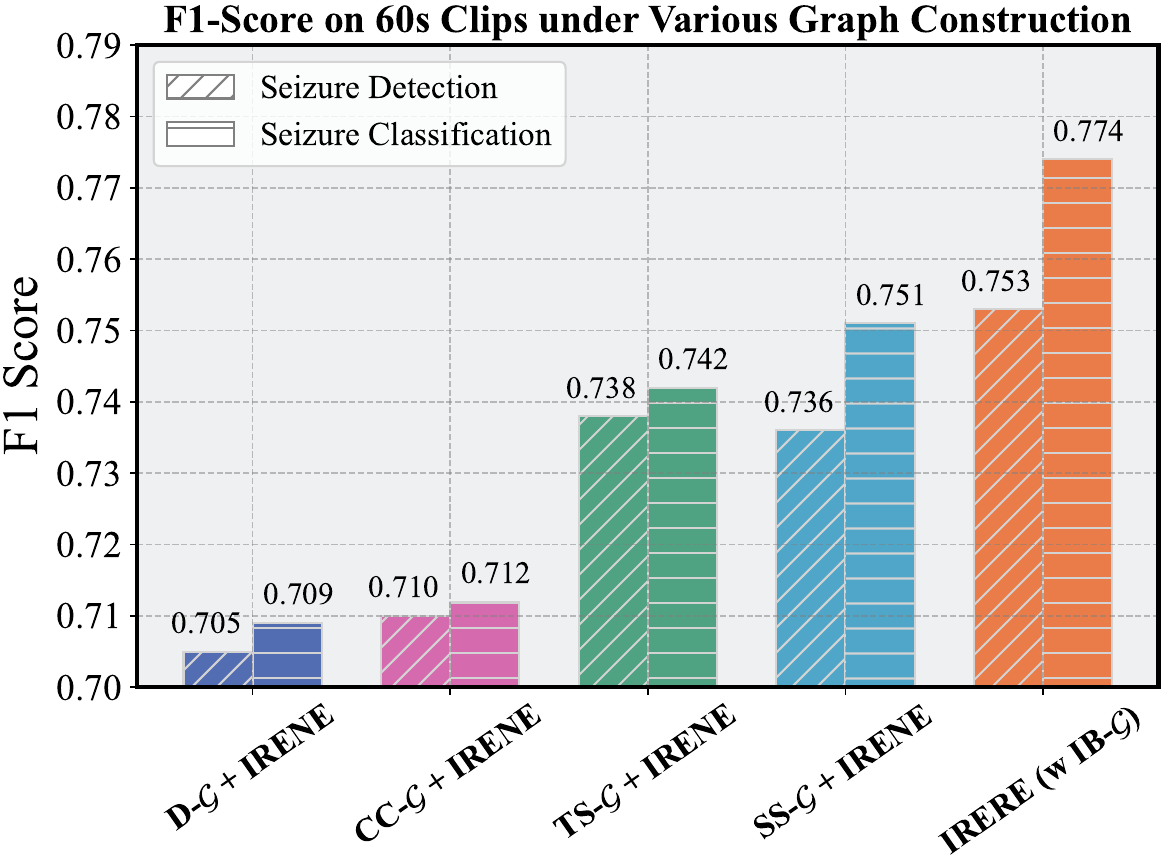}
        \caption{}
        %\caption{F1 across graph construction methods (60s)}
    \end{subfigure}
    \vspace{-2mm}
\caption{Performance of IRENE when using different graph construction methods. We illustrate how various graph structures, including distance-based graph (D-$\mathcal{G}$ + IRENE), cross-correlation (CC-$\mathcal{G}$ + IRENE), temporal similarity-based graph (TS-$\mathcal{G}$ + IRENE), semantic similarity graph (SS-$\mathcal{G}$), and our IB-based dynamic graph, impact the model's F1-Scores.}
\label{fig4-exp-RQ2-graph}
\end{figure}

\begin{figure*}[h]
  \centering
  \includegraphics[width=0.80\textwidth]{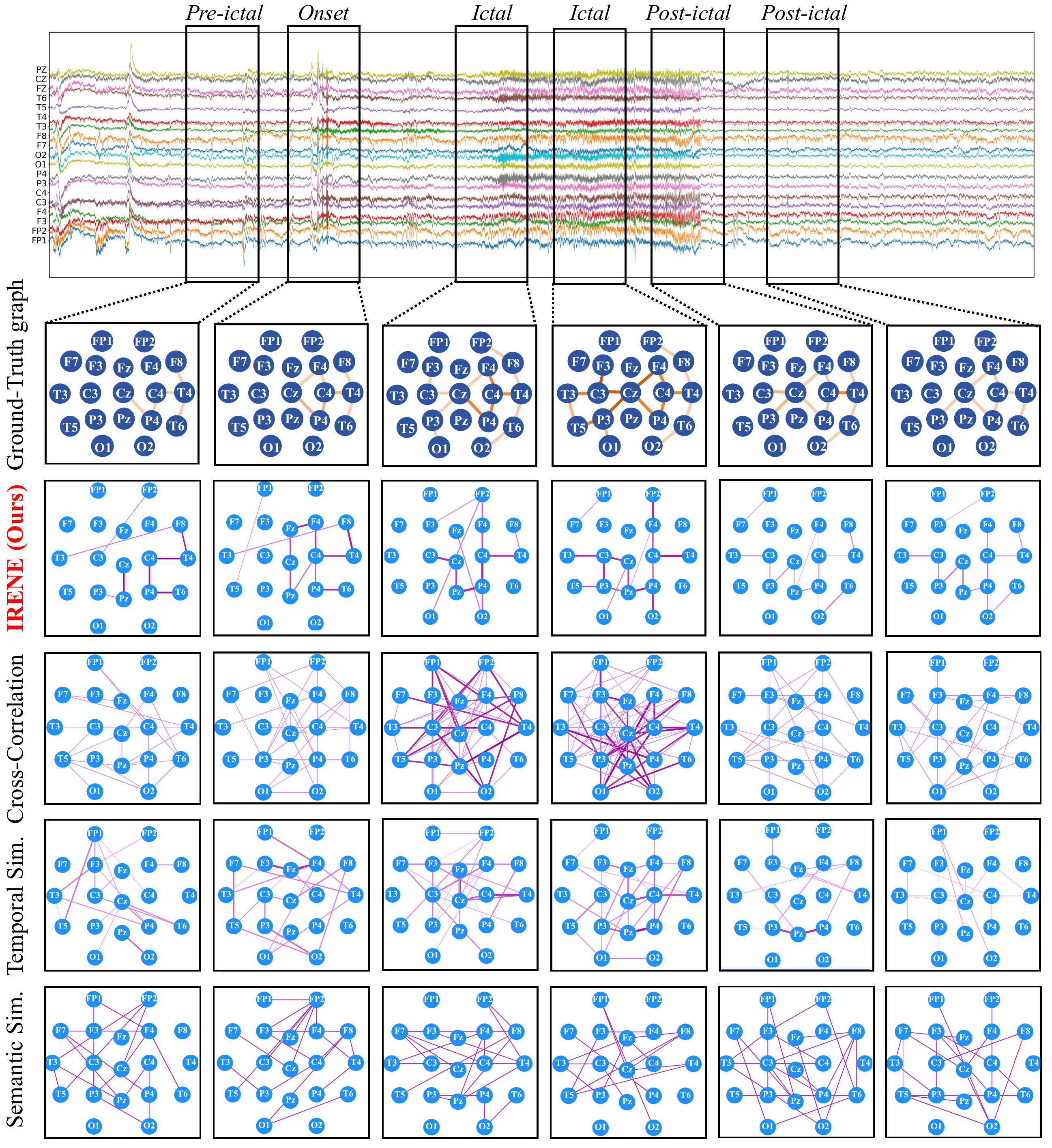}
  \caption{Visualization of learned dynamic EEG graphs across different brain states. Each subfigure in each row shows a learned dynamic graph structure generated by a specific method. We have the following observations for IRENE constructed graphs: In the normal state, the graphs exhibit sparse and weakly connected structures. As seizure onset approaches, the graphs display stronger and more centralized connections, closely resemble to the ground-truth structure patterns and connectivity dynamics.}
  \label{fig5-exp-RQ2-interpretability}
\end{figure*}

To answer \textbf{RQ2}, we evaluate the impact of different graph construction methods on both model performance (Figure~\ref{fig4-exp-RQ2-graph} \& Table~\ref{exp-RQ2-edge-density}) and graph interpretability (Figure~\ref{fig5-exp-RQ2-interpretability}) by comparing IRENE with several mainstream approaches, including Distance-based graph (D-$\mathcal{G}$ + IRENE), Cross-Correlation-based graph (CC-$\mathcal{G}$ + IRENE), Temporal similarity-based (TS-$\mathcal{G}$ + IRENE), and Semantic Similarity-based graph (SS-$\mathcal{G}$). (Figure~\ref{fig4-exp-RQ2-graph} presents the F1 scores of the IRENE model for seizure detection and classification tasks under different graph construction methods, with Figure~\ref{fig4-exp-RQ2-graph} (a) and (b) presents the 12s/60s clips, respectively.) Below, we further explain the labels used in Figure~\ref{fig4-exp-RQ2-graph}. For example, variant ``D-$\mathcal{G}$ + IRENE'' means that we replace the Information Bottleneck-based graph construction method in IRENE with the distance-based graph construction method~\cite{tang2022selfsupervised}. We observe that simple heuristic-based graphs, such as distance-based and cross-correlation-based graphs, yield relatively low F1 scores (e.g., 0.705/0.709 and 0.710/0.712), indicating their limited capability to capture discriminative EEG connectivity patterns. Incorporating more informative similarity measures, such as temporal and semantic similarity, improves performance, reflecting better task-relevant structure modeling. Notably, our proposed IB-based graph construction method achieves the highest F1 scores on detection and classification. This demonstrates the effectiveness of the IB principle in suppressing task-irrelevant noise and redundant edges, leading to more discriminative graph structures and superior downstream performance.

 \vspace{2mm}
 \vspace{-2mm}

\begin{table*}[t]
\centering
\footnotesize
\vspace{-7pt}
\renewcommand{\arraystretch}{0.9}
\caption{Analysis of the contribution of key model components.}\label{tab:ablation}
\resizebox{0.9\linewidth}{!}{
\begin{tabular}{>{\raggedright\arraybackslash}p{2.5cm}|>{\centering\arraybackslash}p{2.5cm}>{\centering\arraybackslash}p{2.5cm}>{\centering\arraybackslash}p{2.5cm}>{\centering\arraybackslash}p{2.5cm}}
\toprule
\multirow{2}{*}{\textbf{Model Variant}} & \multicolumn{2}{c}{\textbf{Detection 60-s}} & \multicolumn{2}{c}{\textbf{Classification 60-s}} \\
\cmidrule(lr){2-3} \cmidrule(lr){4-5}
 & \textbf{F1} & \textbf{AUROC} & \textbf{F1} & \textbf{AUROC} \\
\midrule
w/o IB-$\mathcal{G}$ & 0.728 $\pm$ 0.019 & 0.884 $\pm$ 0.017 & 0.745 $\pm$ 0.018 & 0.896 $\pm$ 0.021 \\
w/o $L_{recon}$ & 0.737 $\pm$ 0.023 & 0.893 $\pm$ 0.022 & 0.758 $\pm$ 0.022 & 0.904 $\pm$ 0.018 \\
w/o GSA-Attn & 0.733 $\pm$ 0.024 & 0.895 $\pm$ 0.026 & 0.754 $\pm$ 0.019 & 0.895 $\pm$ 0.023 \\
w/o Pre-training & 0.721 $\pm$ 0.018 & 0.881 $\pm$ 0.024 & 0.749 $\pm$ 0.023 & 0.893 $\pm$ 0.025 \\
w Full AutoEnc. & 0.732 $\pm$ 0.021 & 0.897 $\pm$ 0.018 & 0.757 $\pm$ 0.016 & 0.901 $\pm$ 0.018 \\
w/o $L_{smooth}$  & 0.744 $\pm$ 0.012 & 0.903 $\pm$ 0.021 & 0.765 $\pm$ 0.013 & 0.908 $\pm$ 0.015 \\
\midrule
\textbf{IRENE}& \textbf{0.753 $\pm$ 0.021} & \textbf{0.916 $\pm$ 0.013} & \textbf{0.774 $\pm$ 0.015} & \textbf{0.918 $\pm$ 0.017} \\
\bottomrule
\end{tabular}
}
\end{table*}

\subsection{Graph Structure Interpretability}
To assess the interpretability of IRENE's learned dynamic graphs for EEG seizure detection, we conduct a qualitative visualization study on a representative EEG sequence where a seizure gradually emerges, intensifies, and eventually subsides. The TUSZ dataset provides channel-level seizure annotations, for example, it specifies which EEG electrodes (e.g., Frontal 7: F7) are involved during seizure events, as identified by clinical experts. Therefore, we select six representative time windows from a seizure recording, spanning the pre-ictal, onset, ictal, and post-ictal phases. For each time window, we visualize the graph structures learned by IRENE and compare them with those generated by several dynamic graph construction baselines, including cross-correlation-based, temporal similarity-based, and semantic similarity-based methods. Additionally, we include the corresponding ground-truth graphs derived from TUSZ annotations to evaluate how effectively each method captures meaningful seizure-related patterns.

As shown in Figure~\ref{fig5-exp-RQ2-interpretability}, in the ground-truth graphs, we observe a clear transition from sparsely connected networks in the pre-ictal phase to more densely connected structures during the ictal period, particularly among the frontal and temporal electrodes.
These stronger and thicker edges indicate clinically relevant synchronization among seizure-related regions. 
Our method, IRENE, captures such transition by adaptively strengthening connections in seizure-critical regions during the ictal phase, while maintaining a sparse and discriminative structure in the pre-ictal and post-ictal phases.
For example, during the ictal phase (column 3-4), IRENE highlights strong edges among electrodes such as F7, T3, and T5, consistent with the ground-truth annotations. In contrast, alternative methods like cross-correlation or temporal similarity often produce overly dense graphs, introducing noisy over-connectivities to the graphs.

\begin{table}[htbp]
\centering
\footnotesize
\caption{Average edge density of each method constructed graphs in Figure~\ref{fig5-exp-RQ2-interpretability}.}
\resizebox{0.70\columnwidth}{!}{
\begin{tabular}{l|c}
%{>{\raggedright\arraybackslash}p{3.3cm}|>{\centering\arraybackslash}p{2.2cm}}
%{l|c}
\toprule
\textbf{Model} & \textbf{Avg. Edge Density} \\
\midrule
Ground-Truth Graph & \underline{0.0575} \\
IRENE (Ours) & \textcolor{red}{0.0636} \\
Cross-Correlation & 0.369 \\
Temporal Similarity & 0.197 \\
Semantic Similarity & 0.183 \\
\bottomrule
\end{tabular}
}
\label{tab:avg_edge_density}
\end{table}

Given another example, during early seizure onset, IRENE emphasizes edges near temporal and central electrodes, which match expert-identified propagation patterns, while others fail to localize such activity. These results highlight that IRENE not only achieves superior  performance but also learns graphs with greater clinical plausibility, supporting its interpretability and potential for aiding neurological diagnosis.

Compared to IRENE, the baseline methods exhibit notable limitations in generating clinically meaningful connectivity patterns. The Cross-Correlation-based graphs often appear overly dense and fail to reflect spatially localized interactions, leading to spurious long-range connections that dilute the signal of seizure-relevant regions. In several windows, this method mistakenly assigns high connectivity to frontal or occipital channels with minimal seizure activity. The Temporal Similarity-based graphs exhibit better temporal smoothness but lack spatial precision, frequently generating uniform or hub-like patterns that obscure localized propagation paths. Semantic Similarity-based graphs encode abstract relationships between EEG channels but struggle to adapt to rapid physiological changes during seizure onset, resulting in graphs that remain largely unchanged across time and fail to capture seizure dynamics. In contrast, IRENE dynamically adjusts its graph structure to emphasize transient, localized interactions aligned with seizure progression. These observations demonstrate that IRENE can capture accurate brain connectivity patterns and offer interpretable graph representations that align more closely with clinically identified epileptogenic zones. The density and regional coverage of learned graphs may provide valuable insights for localizing seizure regions, supporting both surgical planning and fundamental discoveries.

Furthermore, to quantitatively support this observation, we calculate the average edge density for each graph construction method across the 6 selected windows, as shown in Table~\ref{tab:avg_edge_density}. The ground-truth graphs exhibit an average edge density of 0.0575, reflecting their inherently sparse yet clinically meaningful connectivity. IRENE's learned graphs achieve an average density of 0.0636, closely matching the ground truth, indicating its ability to balance sparsity and informativeness. In contrast, the baseline methods generate significantly denser graphs (e.g., Cross-Correlation: 0.369, Temporal Similarity: 0.197, Semantic Similarity: 0.183), underscoring their tendency towards excessive and less meaningful connections.

\subsection{Ablation Study and Component Analysis}

To answer \textbf{RQ3}, we perform ablation studies to assess the contribution of key components in IRENE. Table~\ref{tab:ablation} summarizes the results using simplified variant notations. \textbf{w/o IB-$\mathcal{G}$} removes the IB-based dynamic graph module, causing notable performance drops in both tasks, confirming the necessity of learning sparse, task-relevant inter-channel dependencies. \textbf{w/o $L_{\text{recon}}$}, which discards the masked feature reconstruction loss during pretraining, also degrades performance, demonstrating the role of local context modeling in enhancing representation robustness. \textbf{w/o GSA-Attn} excludes the Graph Structure-Aware Attention, reducing the model capacity to effectively propagate information using graph priors. \textbf{w/o Pretraining}, trained directly on classification without self-supervised pretraining, exhibits the largest performance decline, highlighting the importance of progressive representation learning. \textbf{w Full Autoencoder}, where masked autoencoding is replaced with full autoencoding, shows slight degradation, indicating that selective node masking better encourages the model to focus on informative dependencies. Lastly, \textbf{w/o $L_{\text{smooth}}$} omits the temporal consistency loss, resulting in minor declines yet verifying its role in ensuring stable dynamic connectivity. The results demonstrate that IRENE's superior performance stems from the synergistic integration of IB-guided graph, structure-aware attention, and self-supervised objectives.

\section{Conclusion}
In this work, we present IRENE, a novel structure-aware Graph Masked AutoEncoder framework for EEG-based seizure detection and classification. This work addresses critical challenges in dynamic graph structure learning for noisy and heterogeneous EEG data. Specifically, IRENE introduces an Information Bottleneck-guided graph construction method that explicitly models task-informative and sparse inter-channel dependencies. Furthermore, we design a structure-aware soft mask attention mechanism that leverages structural priors to enhance feature propagation while preserving interpretability. Complemented by a self-supervised masked reconstruction objective, IRENE effectively learns robust and discriminative EEG representations, even under limited labeled data and patient variability. Extensive experiments on real-world seizure datasets demonstrate that IRENE consistently outperforms state-of-the-art baselines across multiple evaluation metrics. IRENE exhibits strong generalization capabilities across varying clip lengths and graph construction paradigms. Overall, this work provides a principled and effective solution for accurate and interpretable seizure diagnosis, with potential future extensions to other graph-based biomedical signal modeling tasks.

% \clearpage

% \section*{ETHICAL CONSIDERATIONS}
% This work uses only publicly available, de-identified EEG datasets collected and shared under appropriate ethical approvals, ensuring that no personally identifiable information is included. The study involves no direct human or animal subjects, and thus does not require additional institutional review. All experiments are conducted in compliance with the terms of use of the datasets. The goal of IRENE is to advance seizure detection methods to assist clinical decision-making and improve patient care, without replacing professional medical judgment. The proposed framework does not make automated clinical diagnoses and should be deployed only in collaboration with qualified healthcare professionals.

\bibliographystyle{IEEEtran}
\bibliography{conference_101719_v2}
\end{document}